\def\year{2020}\relax
\documentclass[letterpaper]{article} % DO NOT CHANGE THIS
\usepackage{aaai20}  % DO NOT CHANGE THIS
\usepackage{times}  % DO NOT CHANGE THIS
\usepackage{helvet} % DO NOT CHANGE THIS
\usepackage{courier}  % DO NOT CHANGE THIS
\usepackage[hyphens]{url}  % DO NOT CHANGE THIS
\usepackage{graphicx} % DO NOT CHANGE THIS
\usepackage{graphicx}  % DO NOT CHANGE THIS
\usepackage{diagbox}
\usepackage{slashbox}
\usepackage{amsmath}
\usepackage{amssymb}
\usepackage{subfigure}
\usepackage{multirow}
\usepackage{nidanfloat}
\usepackage{comment}
\usepackage{xcolor}
\title{Single Camera Training for Person Re-identification}

\author{Tianyu Zhang,\textsuperscript{\rm 1} Lingxi Xie,\textsuperscript{\rm 2} Longhui Wei,\textsuperscript{\rm 3} Yongfei Zhang,\textsuperscript{\rm 1} Bo Li,\textsuperscript{\rm 1} Qi Tian \textsuperscript{\rm 4}\\
\textsuperscript{\rm 1}School of Computer Science and Engineering, Beihang University, \textsuperscript{\rm 2}Johns Hopkins University,\\
\textsuperscript{\rm 3}Peking University, \textsuperscript{\rm 4}School of Electronic Engineering, Xidian University\\
zhangtianyu@buaa.edu.cn, 198808xc@gmail.com, longhuiwei@pku.edu.cn, \\
yfzhang@buaa.edu.cn, boli@buaa.edu.cn, wywqtian@gmail.com
}
\begin{document}

\maketitle

\begin{abstract}
Person re-identification (ReID) aims at finding the same person in different cameras.
Training such systems usually requires a large amount of cross-camera pedestrians to be annotated from surveillance videos, which is labor-consuming especially when the number of cameras is large.
Differently, this paper investigates ReID in an unexplored single-camera-training (SCT) setting, where each person in the training set appears in only one camera.
To the best of our knowledge, this setting was never studied before.
SCT enjoys the advantage of low-cost data collection and annotation, and thus eases ReID systems to be trained in a brand new environment.
However, it raises major challenges due to the lack of cross-camera person occurrences, which conventional approaches heavily rely on to extract discriminative features.
The key to dealing with the challenges in the SCT setting lies in designing an effective mechanism to complement cross-camera annotation.
We start with a regular deep network for feature extraction, upon which we propose a novel loss function named multi-camera negative loss (MCNL).
This is a metric learning loss motivated by probability, suggesting that in a multi-camera system, one image is more likely to be closer to the most similar negative sample in other cameras than to the most similar negative sample in the same camera.
In experiments, MCNL significantly boosts ReID accuracy in the SCT setting, which paves the way of fast deployment of ReID systems with good performance on new target scenes.
\end{abstract}

\section{Introduction}

Person re-identification (ReID) aims to retrieve a certain person appearing in a camera network.
With increasing concerns on public security, ReID has attracted more and more research attention from both academia and industry.
In the past years, many algorithms~\cite{Sun2018_beyond,streid,suh_eccv18,PAN} and datasets~\cite{MARKET,DUKEMTMC,PTGAN,RPIfield} have been proposed, which significantly boosted the progress of this research field.
Despite the higher and higher accuracy obtained by specifically designed approaches on standard ReID benchmarks, many issues of this task remain unsolved.
As discussed in~\cite{PTGAN,SPGAN}, a ReID model trained on one dataset performs poorly on other datasets due to dataset bias.
Thus, to deploy a ReID system to a new environment, labelers have to annotate a training dataset from the target scene, which is often time-consuming and even impractical in large-scale application scenarios.
To tackle this issue, researchers make a common assumption is that there exist a number of available unlabeled images in the target scene, based on which they design some unsupervised learning~\cite{Yutian2019}
or domain adaptation approaches~\cite{TJAIDL} to improve ReID performance.
However, these methods depend on extra modules to predict the pseudo label of each image or generate fake images.
Therefore, it is not always reliable and reports less satisfying performance compared to supervised learning methods.

\begin{figure}
  \centering
  \includegraphics[width=0.46\textwidth]{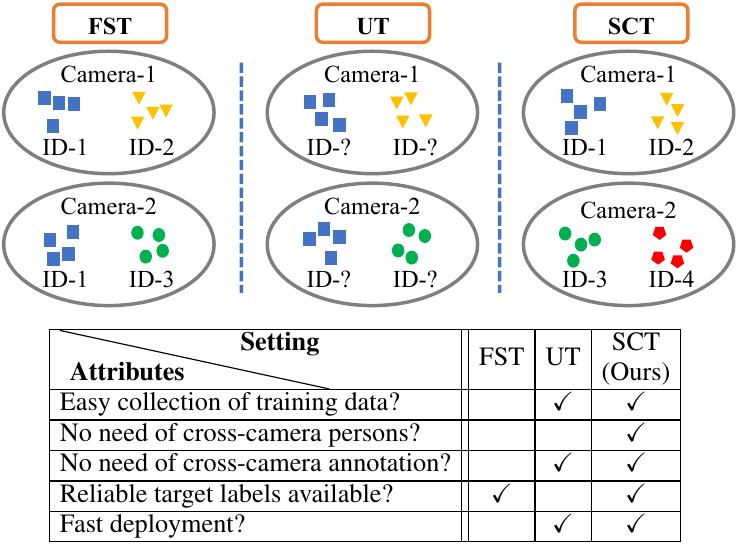}
  \caption{The comparison between SCT and previous settings in person ReID. \textit{Fully-supervised-training} (FST) data are composed of annotated pedestrians appearing in multiple cameras. \textit{Unsupervised-training} (UT) data have no identity annotations. Under our \textit{single-camera-training} (SCT) setting, each pedestrian appears in only one camera and identity labels are easy to obtain.}
  \label{fig:fig1}
\end{figure}

Different from previous work, this paper investigates ReID under a novel single-camera-training (SCT) setting, where each pedestrian appears in only one camera.
We compare our setting to previous ones in Fig.~\ref{fig:fig1}.
Without the heavy burden of annotating cross-camera pedestrians, training data with labels are easy to obtain under SCT.
For example, using off-the-shelf tracking techniques~\cite{tracking,tracking2}, researchers can quickly collect a large number of tracklets under each camera at different time periods, and thus it is very likely that each of them corresponds to a unique ID.
Therefore, compared to the fully-supervised-training (FST) setting, {\em i.e.}, learning knowledge from cross-camera annotations, SCT requires much fewer efforts in preparing for training data.
Compared to the unsupervised-training (UT) setting, which requires frequent cross-camera person occurrences, SCT makes a mild assumption on camera-independence, so as to provide weak but reliable supervision signals for learning.
Therefore, SCT has the potential of being deployed to a wider range of application scenarios.

It remains an issue of how to make use of the camera-independence assumption to learn discriminative features for ReID.
The most important one lies in camera isolation, which implies that there are no cross-camera pedestrians in the entire training set.
Conventional methods heavily rely on cross-camera annotations because this is the key supervision that a model receives for metric learning.
That is to say, by pulling the images of the same person appearing in different cameras close, conventional methods can learn camera-unrelated features so that they perform well on the testing set.
With camera isolation in SCT, we must turn to other types of supervision to achieve the goal of metric learning.

To this end, we propose a novel loss term named Multi-Camera Negative Loss (MCNL).
The design of MCNL is inspired by a simple hypothesis, that given an arbitrary person in a multi-camera network, it is more likely that the most similar person is found in another camera, rather than in the same camera, because there are simply more candidates in other cameras.
To verify this, we perform statistical analysis on several public datasets, and the results indeed support our assumption (please see Fig.~\ref{fig:prob}).
Based on the above observation, our MCNL adjusts feature distributions and alleviates camera isolation problem by ranking the distances of cross-camera negative pairs and within-camera negative pairs.
Extensive experiments show that MCNL can force the backbone network to learn more person-related features but ignore camera-unrelated representations, and then achieves good performance under the SCT setting.

Our major contributions can be summarized as follows:
\begin{itemize}
    \item
    To the best of our knowledge, this paper is the first to present the SCT setting. Moreover, this paper analyzes the advantages and challenges under the SCT setting compared to existing settings in person ReID.
    \item
    To solve the issue of camera isolation under the SCT setting, this paper proposes a simple yet effective loss term named MCNL. Extensive experiments show that MCNL significantly boosts the ReID performance under SCT, and it is not sensitive to wrong annotations.
    \item
    Last but not least, by solving SCT, this paper sheds light on fast deployment of ReID systems in new environments, implying a wide range of real-world applications.
\end{itemize}

\section{Related Work}
\label{sec:2}

Our work is proposed under the new single-camera-training setting, which is relative to previous FST setting and UT setting.
In this section, we mainly summarize the existing methods of these settings and then elaborate the differences between these settings and our SCT.

\subsection{Fully-Supervised-Training Setting}
The FST setting implies that there are a large number of annotated cross-camera pedestrian images for training.
In previous works under the FST setting, most of them formulated person ReID as a classification task and trained a classification model with the labeled training data~\cite{zheng2016person,zheng2016a,SVDNET}.
With the advantages of large-scale training data and deep neural networks, these methods achieve good results.
In addition, some researchers designed complex network architectures to extract more robust and discriminative features~\cite{GLAD,alignedreid,POSETRANS}.
Differently, other researchers argue that the surrogate loss for classification may not be suitable when the number of identities increases~\cite{InDefense}.
Therefore, end-to-end deep metric learning methods were proposed and widely used under the FST setting~\cite{centerloss,InDefense,Quadruplet}.
For example, Hermans~\emph{et~al.}~\cite{InDefense} demonstrated that the triplet loss is more effective for person ReID task.
Chen~\emph{et~al.}~\cite{Quadruplet} proposed a deep quadruplet network to improve the ReID performance further.
Although the performance has been boosted significantly, the demand for annotating large-scale training data hinders their real-world applications,~\emph{e.g.}, the fast deployment of ReID systems in new target scenes is almost impossible.
This is because it is rather expensive to collect this kind of training data for the FST setting.
Different from the FST setting, the SCT setting requires much less time in the training data collection process, since there is no need to collect and annotate cross-camera pedestrian images.
Therefore, our SCT setting is more suitable for fast deployment of ReID systems in new target scenes.

\subsection{Unsupervised-Training Setting}
Different from FST, the UT setting means there are no labeled training data.
Although hand-crafted features like LOMO~\cite{LOMO}, BOW~\cite{MARKET} and ELF~\cite{ELF} can be used directly, the ReID performance is relatively low.
Therefore, some researchers designed novel unsupervised learning methods to improve ReID performance under the UT setting.
Liang~\emph{et~al.}~\cite{Liang2015} proposed a salience weighted model.
Lin~\emph{et~al.}~\cite{Yutian2019} adopted a bottom-up clustering approach for purely unsupervised ReID.
Without the supervision of identity labels, the performance of their methods is still not satisfactory.
To further boost ReID accuracy, many unsupervised domain adaptation methods have been proposed.
They conducted supervised learning on the source domain and transferred to the target domain, thus can benefit from FST and produce better results.
The ways of transferring domain knowledge include image-image translation~\cite{PTGAN,SPGAN}, attribute consistency scheme~\cite{TJAIDL}, and so on~\cite{HHL,peng2016,ECN}.
These methods perform well when the target domain and the source domain are very similar, but may not be suitable when the domain gap is large~\cite{TAUDL}.
Differently, the problem does not exist in our proposed SCT setting because under SCT, ReID models are only trained with training data from the target scene.
More recently, Li~\emph{et~al.}~\cite{TAUDL} built the cross-camera tracklet association to learn a robust ReID model from automatically generated person tracklets.
This method~\cite{TAUDL} assumes that cross-camera pedestrians are common, and thus camera relations can be learned by matching person tracklets.
However, in a large-scale camera network, the average number of cameras pedestrians pass through is quite small,~\emph{e.g.}, one person appears in only five cameras from thousands of cameras.
Moreover, the tracklet association method~\cite{TAUDL} is not so reliable to make sure each matched tracklets belonging to the same person.
The wrong matched tracklets may cause the learned ReID model to perform poorly. Inspired by the above discussion, we propose a more reliable setting,~\emph{i.e.},single-camera-training, and further design the multi-camera negative loss to improve the ReID performance under this setting.

\section{Problem: Single-Camera-Training}
Researchers report major difficulty in collecting and annotating data for person ReID, and such difficulty is positively related to the number of cameras in the network. We take MSMT17~\cite{PTGAN}, a large-scale ReID dataset, as an example. To construct it, researchers collected high-resolution videos covering 180 hours from 15 cameras, after which three labelers worked on the data for 2 months for cross-camera annotation. In another dataset named RPIfield~\cite{RPIfield}, there are two types of pedestrians known as actors and distractors, respectively. A small number of actors followed pre-defined paths to walk in the camera network, and thus it is easy to associate the images captured by different cameras. However, a large number of distractors, without being controlled, are walking randomly, so that it is rather expensive to annotate these pedestrians among cameras. This annotation process, from a side view, verifies the difficulty of collecting and annotating cross-camera pedestrians.

On the other hand, cross-camera information plays the central role in person ReID, because for the conventional approaches, this is the main source of supervision that the same person appears differently in the camera network -- this is exactly what we hope to learn. We quantify how existing datasets provide cross-camera information by computing the average number of occurrences of each person in the camera network,~{\em i.e.}, if a person appears in three cameras, his/her number of occurrences is 3. We name it the camera-per-person (CP) value and list a few examples in Tab.~\ref{tab:cp}. We desire a perfect dataset in which all persons are annotated in all cameras,~{\em i.e.}, CP equals to the number of cameras, but for a large camera network, this is often impossible,~{\em e.g.}, in MSMT17, the CP value is 3.81, far smaller than 15, the number of cameras.

\begin{table}
    \centering
      \caption{The camera-per-person (CP) value of a few ReID datasets. $N_\mathrm{cam}$ denotes the number of cameras.}
    \begin{tabular}{l||c|c|c}
        \hline
        Dataset & $N_\mathrm{cam}$ & $\mathrm{CP}$ & $\mathrm{CP}/N_\mathrm{cam}$ \\
        \hline\hline
        MSMT17 & 15 & 3.81 & 0.254 \\
        DukeMTMC-reID          & 8 &  3.13 & 0.391\\
        Market-1501            & 6 &  4.34  & 0.724 \\
        RPIfield (distractors) & 12 &  1.25 &  0.104 \\
        RPIfield (actors)      & 12 &  6.99  & 0.583 \\
        RPIfield (total)       & 12 &  1.40  & 0.117 \\
        \hline
    \end{tabular}
    \label{tab:cp}
\end{table}

To alleviate the burden of data annotation, we propose to consider the scenario that no cross-camera annotations are available,~{\em i.e.}, CP equals to 1 regardless of the number of cameras in the network. We name this setting to be {\bf single-camera-training} (SCT)\footnote{{\em Here is a disclaimer}: SCT does not mean that there is only one single camera in the training data, but we simply assume that each person appears in only one camera, or, in other words, two persons appearing in different cameras must have different identities.}.
This requirement can be achieved by collecting data from different cameras in different time periods ({\em e.g.}, recording camera A from 8 am to 9 am while camera B from 10 am to 11 am) -- although this cannot guarantee our assumption, as we shall see in experiments, our approach is robust to a small fraction of `outliers',~{\em i.e.}, two or more occurrences of the same person in different cameras are assumed to be different identities.

This setting greatly eases the fast deployment of a ReID system. With off-the-shelf person tracking algorithms~\cite{tracking,tracking2}, we can easily extract a large number of {\em tracklets} in videos, each of which forms an identity in the training set. However, such a training dataset is less powerful than those specifically designed for the ReID task, as it lacks supervision of how a person can appear differently in different cameras. We call this challenge {\bf camera isolation}, and will elaborate on this point carefully in the next section.

\section{Our Approach}

\subsection{Baseline and Motivation} \label{sec:4.1}

Existing ReID approaches often start with a backbone which extracts a feature vector $\mathbf{x}_k$ from an input image $\mathbf{I}_k$. On top of these features, there are mainly two types of loss functions, and sometimes they are used together towards higher accuracy. The first type is named the {\em cross entropy} (CE) loss, which requires the model to perform a classification task in which the same person in different cameras are categorized into one class. The second type is named the {\em triplet margin} (TM) loss, which assumes that the largest distance between two appearances of the same person should be smaller than the smallest distance between this person and another. When built upon a ResNet-50~\cite{ResNet} backbone, CE and TM achieve 78.9\% and 79.0\% rank-1 accuracy on the DukeMTMC-reID dataset~\cite{DUKEMTMC}, respectively, without any bells and whistles.

However, in the SCT setting, both of them fail dramatically due to camera isolation. From the DukeMTMC-reID dataset, we sample 5,993 images from the training set which satisfy the SCT setting, and the corresponding models, with CE and TM losses, report 40.2\% and 21.2\%, respectively. In comparison, we sample another training subset with the same number of images but equipped with cross-camera annotation, and these numbers become 69.3\% and 75.8\%, which verifies our hypothesis.

\begin{figure}[!t]
    \centering
    \includegraphics[width=0.47\textwidth]{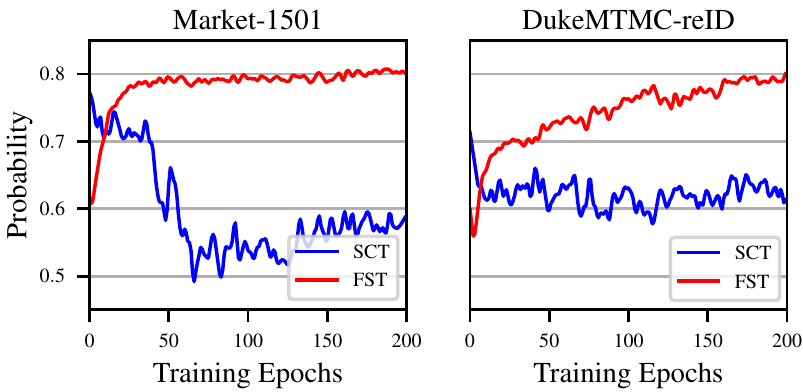}
    \caption{Curves of the probability produced by the triplet margin loss, with respect to the number of elapsed training epochs, of finding the most similar person (of a different ID) in another camera. The figures on the left and right show results on Market-1501 and DukeMTMC-reID, respectively.}
    \label{fig:prob}
\end{figure}

To explain this dramatic accuracy drop, we first point out that a ReID system needs to learn feature embedding which is independent to cameras ({\em i.e.}, camera-unrelated features), which is to say, the learned feature distribution is approximately the same under different cameras. However, we point out that both CE and TM losses cannot achieve this goal by themselves -- they heavily rely on cross-camera annotations. Without such annotations, existing ReID systems often learn camera-related features.

Here, we provide another metric to quantify the impact of camera-related/unrelated features, which is the core observation that motivates our algorithm design. Intuitively, for a set of camera-unrelated features, the feature distribution over the entire camera network should be approximately the same as the distribution over any single camera. In other words, the expectation of similarity between two different persons in the same camera should {\bf not} be higher than that between two different persons from two cameras. Therefore, given an anchor image, the probability that the most similar person appears in the same camera is only $1/N_\mathrm{cam}$,~{\em i.e.}, in a multi-camera system, the most similar person mostly appears in another camera. Thus, we perform statistics during the training process with the TM loss, under both SCT and FST, and show results in Fig.~\ref{fig:prob}. We can see that, under the FST setting, this probability is mostly increasing during the training process, and eventually reaches a plateau at around 0.8; while under the SCT setting, the curve is less stable and the stabilized probability is much lower.

Thus, our motivation is to facilitate the learned features to satisfy that the most similar person appears in another camera. {\bf This is considered as the extra, weakly-supervised cue to be explored in the SCT setting.} This leads to a novel loss function, the Multi-Camera Negative Loss (MCNL), which is detailed in the next subsection.

\subsection{Multi-Camera Negative Loss}
Inspired by the analyses above, we design the Multi-Camera Negative Loss (MCNL) to ensure that, given any anchor image in one camera, the most similar negative image is more likely to be found from other cameras, and the negative image should be less similar to the anchor image, compared to the most dissimilar positive image.

In a mini-batch with $C$ cameras, $P$ identities from each camera and $K$ images of each identity ({\em i.e.}, the batch size is $C\times P\times K$), given an anchor image $\mathbf{I}_k^{c,p}$, let $f_\theta(\mathbf{I}_{k}^{c,p})$ denote the feature mapping function learned by our network, and $\|f_1-f_2\|$ represent the Euclidean distance between two feature vectors. The hardest positive distance of $\mathbf{I}_k^{c,p}$ is defined as:
\begin{equation}
    \mathrm{dist}_{+}^{c,p,k}=\max_{l=1...K,l\ne k}\|f_\theta(\mathbf{I}_k^{c,p})-f_\theta(\mathbf{I}_l^{c,p})\|.
\end{equation}
Then, we have the hardest negative distance in the same camera:
\begin{equation}
    \mathrm{dist}_{-,\mathrm{same}}^{c,p,k}=\min_{\substack{l=1...K,\\q=1...P,q\ne p}}\|f_\theta(\mathbf{I}_k^{c,p})-f_\theta(\mathbf{I}_l^{c,q})\|,
\end{equation}
and the hardest negative distance in other cameras:
\begin{equation}
    \mathrm{dist}_{-,\mathrm{other}}^{c,p,k}=\min_{\substack{l=1...K,\\q=1...P,\\ o=1...C, o\ne c}}\|(f_\theta(\mathbf{I}_k^{c,p})-f_\theta(\mathbf{I}_l^{o,q})\|.
\end{equation}
With these terms, MCNL is formulated as follows:
\begin{align}\label{eq:L}
    \mathcal{L}_\mathrm{MCNL}=& \sum_{c=1}^C\sum_{p=1}^P\sum_{k=1}^K [m_1+ \mathrm{dist}_{+}^{c,p,k} -\mathrm{dist}_{-,\mathrm{other}}^{c,p,k}]_+ \nonumber \\
  &  +[m_2+\mathrm{dist}_{-,\mathrm{other}}^{c,p,k}-\mathrm{dist}_{-,\mathrm{same}}^{c,p,k}]_+,
\end{align}
where $[z]_+=\max(z,0)$, and both of $m_1$ and $m_2$ denote the values of margins.

As shown in Eq.~\eqref{eq:L}, the second loss term is to ensure the most similar negative image is found from other cameras, and the first loss term is to force this negative image to be less similar than the most dissimilar positive image.
These two parts together provide boundaries to restrict $\mathrm{dist}_{-,\mathrm{other}}^{c,p,k}$ between $\mathrm{dist}_{+}^{c,p,k}$ and $\mathrm{dist}_{-,\mathrm{same}}^{c,p,k}$, which meets the motivation described in the previous section.

Moreover, the proposed MCNL also ensures that the learned feature is discriminative and camera-unrelated.
Given the most similar cross-camera negative image differs in camera factors, it is more likely that the similarity lies in person appearance.
By pulling the most similar cross-camera negative pairs a little closer, MCNL encourages the model to focus more on person appearance.
For the most similar within-camera negative pair, as camera factors are shared with the anchors, pushing them away further reduces the impact of cameras. % and focuses on the difference from foreground.
In addition, MCNL also ensures positive pairs closer than cross-camera negative pairs, which meets the basic correctness of metric learning.

\textbf{Differences from prior work.}
Previously, researchers proposed many triplet-based or quadruplet-based loss functions to improve ReID performance~\cite{InDefense,FACENET,Moderate}.
The largest difference between our approach and theirs lies in that they pushed away the hardest negative images from other cameras without constraints, while we do not. In a dataset constructed under the SCT setting, these methods tend to learn camera-related cues to separate negative images from another camera, which further aggravates the camera isolation problem.
Moreover, we evaluate several state-of-the-art methods related to metric learning and ReID under SCT. The experiment results demonstrate that existing methods are not suitable for this new setting (please see Sec.~\ref{sec:5.4}).

\textbf{Advantages.}
Based on the above discussions, the advantages of our proposed MCNL can be summarized as two folds.
(\romannumeral1)  MCNL can alleviate the camera isolation problem. Through pulling the cross-camera negative pairs closer and pushing the within-camera negative pairs away, MCNL forces the feature extraction model to ignore the camera clues.
(\romannumeral2) Same with previous metric learning approaches,  MCNL can force the feature extraction model to learn a more discriminative representation by adding the constraint that, the hardest positive image should be closer to the anchor image, compared with the negative images (both cross-camera and within-camera negative images).

\section{Experiments}
\subsection{Datasets} \label{sec:5.1}

\begin{table}
    \caption{Detailed statistics of the datasets used in our experiments.}
    \small
    \centering
    \setlength{\tabcolsep}{0.08cm}
    \begin{tabular}{l|c|c|c|c|c}
        \hline
        Dataset & \begin{tabular}[c]{@{}c@{}}\#Train\\IDs\end{tabular} & \begin{tabular}[c]{@{}c@{}}\#Train\\Images\end{tabular}  & \begin{tabular}[c]{@{}c@{}}\#Test\\IDs\end{tabular}  &
        \begin{tabular}[c]{@{}c@{}}\#Test\\Images\end{tabular} &                 \begin{tabular}[c]{@{}c@{}}With cross-\\camera persons?\end{tabular}  \\
        \hline\hline
        Market  & 751 & 12,936 & 750 & 15,913 & True \\
        \hline
        Market-SCT  & 751 & 3,561 & 750 & 15,913 & False \\
        \hline
        Duke &  702 & 16,522 & 1,110 & 17,661 & True \\
        \hline
        Duke-SCT & 702 & 5,993 & 1,110 & 17,661 & False \\
        \hline
    \end{tabular}
    \label{tab:datasets}
\end{table}

To evaluate the effectiveness of our proposed method, we mainly conduct experiments on two large-scale person ReID datasets,~\emph{i.e.}, Market-1501~\cite{MARKET} and DukeMTMC-reID~\cite{DUKEMTMC}.
For short, we refer to Market-1501 and DukeMTMC-reID as Market and Duke, respectively.

Both Market and Duke are widely used person ReID datasets.
For each person in the training sets, there are multiple images from different cameras.
To better evaluate our method, we reconstruct these training sets for the SCT setting.
More specifically, we randomly choose one camera for each person and take those images of the person under the selected camera as training images.
Finally, we sample 5,993 images from the training set of Duke and 3,561 images from the training set of Market.
In this paper, we denote these sampled datasets as Duke-SCT and Market-SCT, respectively.
Note that, we still keep the original testing data and strictly follow the standard testing protocols.
The detailed statistics of the datasets are shown in Tab.~\ref{tab:datasets}.

\subsection{Implementation Details}
We adopt ResNet-50~\cite{ResNet} which is pre-trained on ImageNet~\cite{ImageNet} as our network backbone.
The final fully connected layers are removed, and we conduct global averaging pooling (GAP) to the output of the fourth block of ResNet-50.
The 2048-dim GAP feature is used for metric learning.
In each batch, we randomly select $8$ cameras, and sample $4$ identities for each selected camera.
Then, we randomly sample $8$ images for each identity, leading to the batch size of $256$ for Duke-SCT.
For Market-SCT, there are only $6$ cameras in the training set.
Hence, we sample $6$ cameras, $5$ identities for each camera, and $8$ images for each identity, thus the batch size is $240$ for Market-SCT.
We empirically set $m_1$ and $m_2$ as $0.1$, respectively.
For baseline, we implement the batch hard triplet loss~\cite{InDefense}, which is one of the most effective implementations of the TM loss.
For short, we use Triplet to denote the batch hard triplet loss in the following sections.
The margin of Triplet is set to be $0.3$, as it achieves excellent performance under the FST setting.
The input images are resized as $256\times128$, and Adam~\cite{ADAM} optimizer is adopted.
Weight decay is set as $5\times10^{-4}$.
The learning rate $\epsilon$ is initialized as $2\times10^{-4}$ and exponentially decays following the Eq.~\eqref{eq:lr} proposed in~\cite{InDefense}:
\begin{equation}
\epsilon(t)=\left\{
             \begin{array}{ll}
             \epsilon_0, & t\leq t_0 \\
             \epsilon_0 \times0.001^{\frac{t-t_0}{t_1-t_0}}, & t_0 \leq t \leq t_1.
             \end{array}
\right.
\label{eq:lr}
\end{equation}
For all datasets, we update the learning rate every epoch after 100 epochs and stop training when reaching 200 epochs,~\emph{i.e.}, $t_0=100$ and $t_1=200$, respectively.
All experiments are conducted on two NVIDIA GTX 1080Ti GPUs.

\subsection{Diagnostic Studies} \label{sec:5.3}

\begin{table}[]
    \centering
    \caption{ReID accuracy (\%) produced by different loss terms, among which MCNL reports the best results. The training sets are Duke-SCT and Market-SCT, respectively. Triplet, Triplet-other and Triplet-same denote the batch hard triplet loss~\cite{InDefense} and its two variations, respectively.}

    \begin{tabular}{l|cc|cc}
        \hline
        \multirow{2}*{Methods} & \multicolumn{2}{c|}{ Duke-SCT } & \multicolumn{2}{c}{ Market-SCT } \\
        \cline{2-5}
         & Rank-1 & mAP & Rank-1 & mAP \\
        \hline\hline
        Triplet & 21.2 & 11.3 & 39.7 & 18.2\\
        \hline
        Triplet-other & 9.9 & 3.6 & 25.2 & 8.8\\
        \hline
        Triplet-same & 54.6  & 35.9 & 51.3 & 28.0\\
        \hline
        MCNL & \textbf{66.4} & \textbf{45.3} & \textbf{66.2} & \textbf{40.6} \\
        \hline
    \end{tabular}
    \label{tab:ana}
\end{table}

\textbf{The effectiveness of MCNL.}
MCNL is designed based on Triplet~\cite{InDefense}.
To better demonstrate the effectiveness of MCNL, we evaluate the performance of Triplet and its two variations, Triplet-same and Triplet-other.
Triplet-same represents the hardest negative image is selected from the same camera as the anchor image, and Triplet-other means the hardest negative image is found from other cameras.
The performance of these methods is summarized in Tab.~\ref{tab:ana}.

\begin{figure}[!t]
    \centering
    \subfigure[Triplet (8.433)]{
        \includegraphics[width=3.5cm]{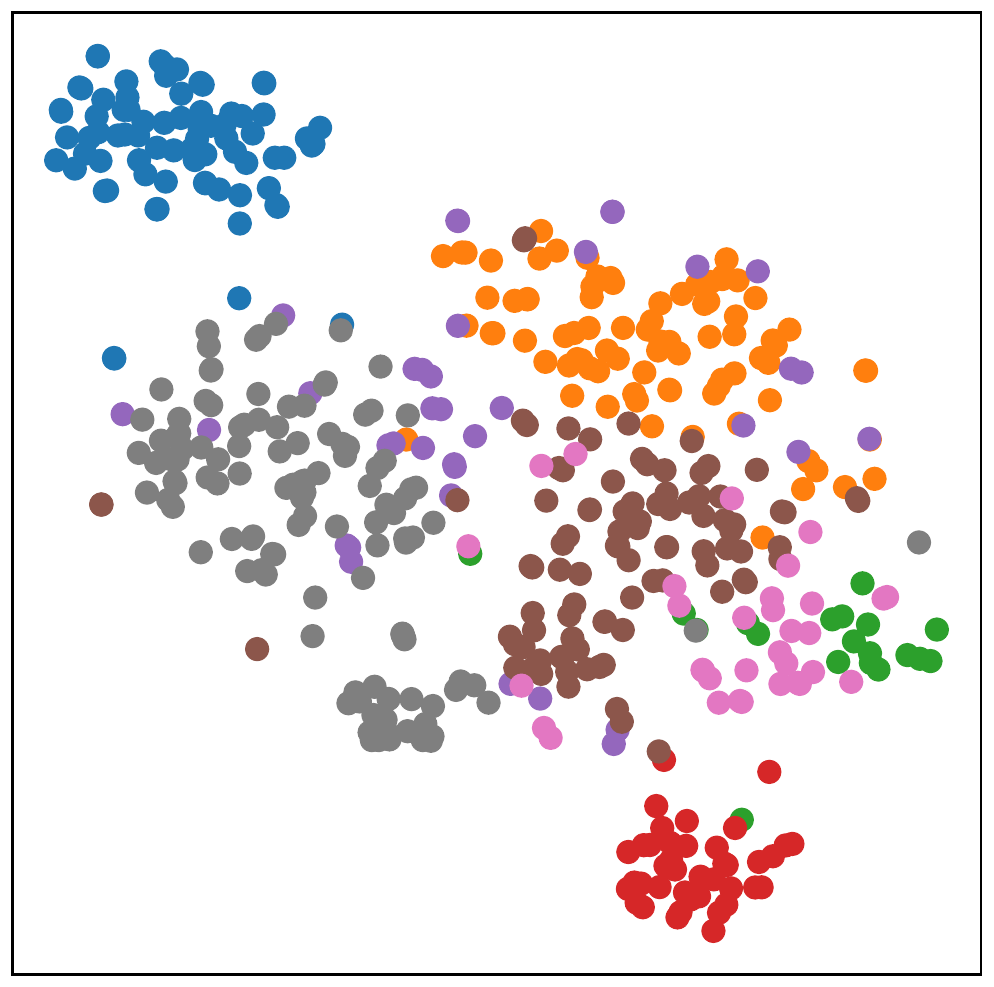}
    }
    \quad
    \subfigure[Triplet-other (16.683)]{
        \includegraphics[width=3.5cm]{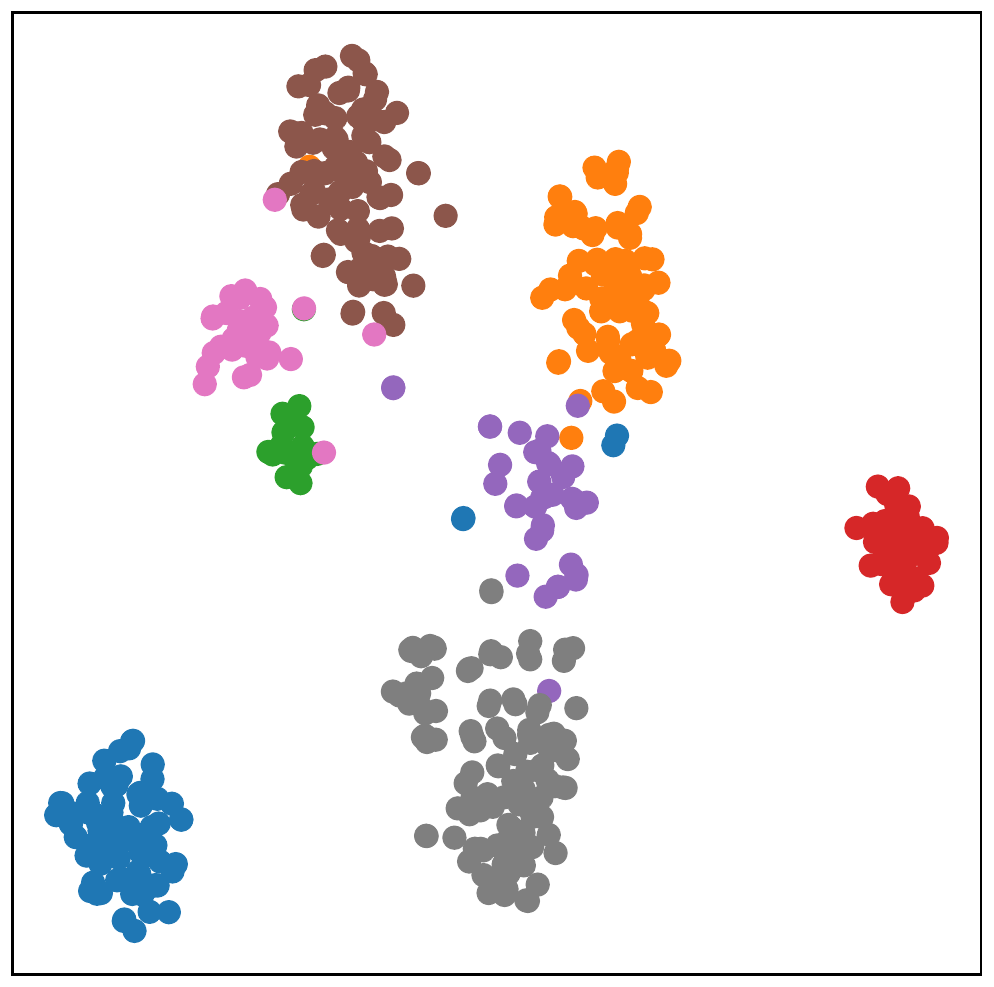}
    }
    \quad
    \subfigure[Triplet-same (0.404)]{
        \includegraphics[width=3.5cm]{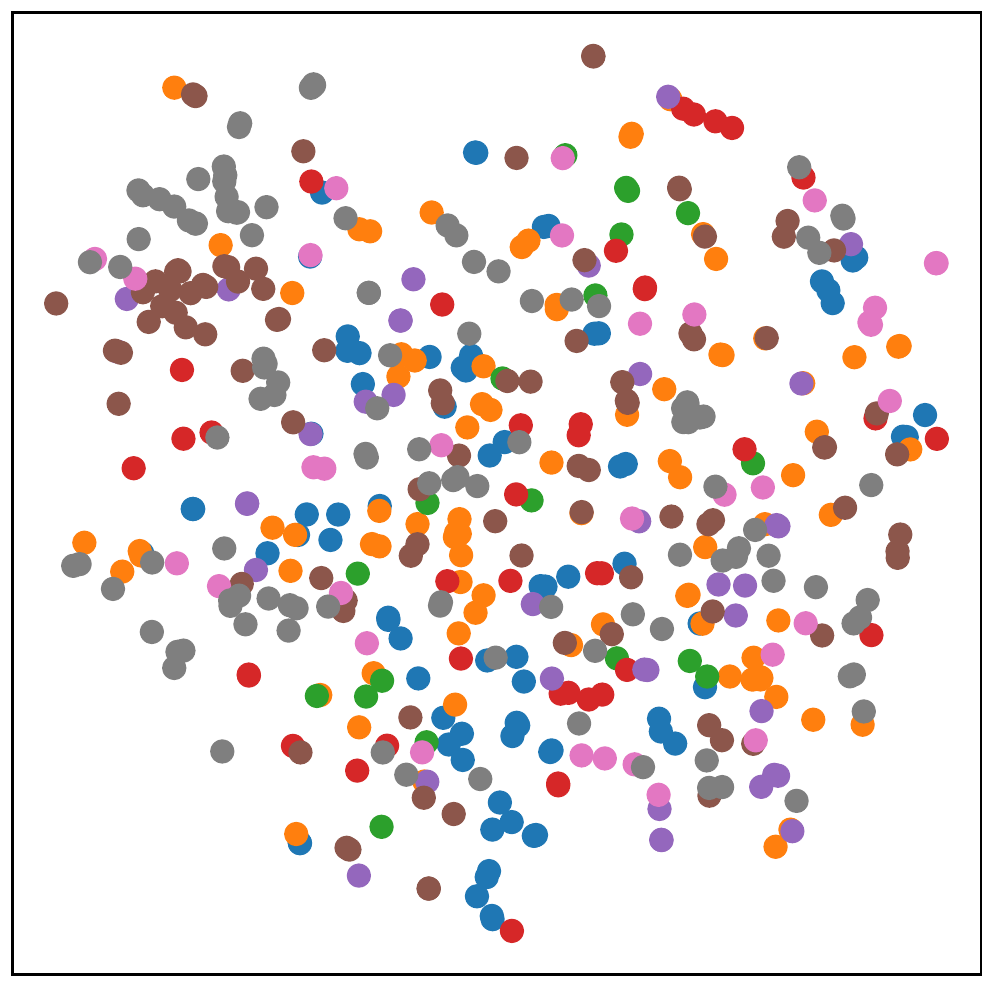}
    }
    \quad
    \subfigure[MCNL (0.255)]{
        \includegraphics[width=3.5cm]{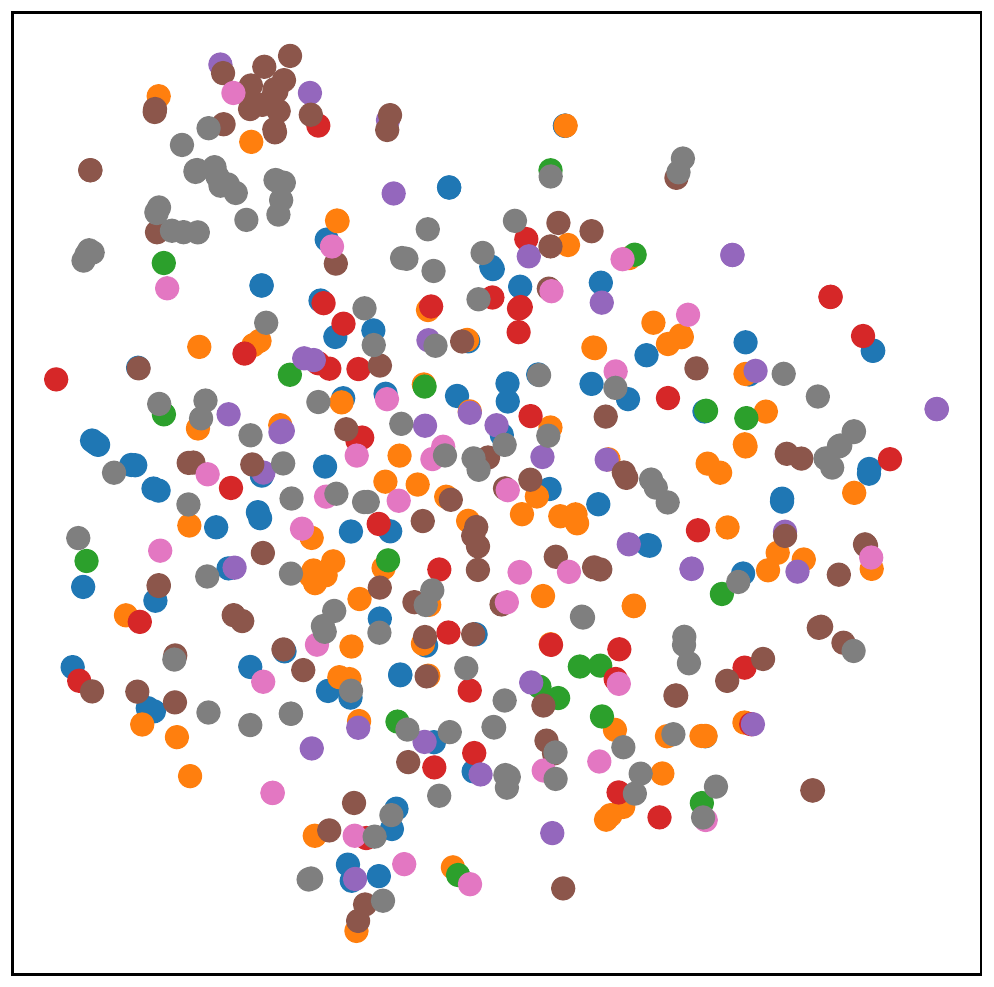}
    }
    \caption{Visualization of feature distributions. Pseudo \textit{F} statistics are shown in brackets. Each color indicates features from a camera. This figure is best viewed in color. }
    \label{fig:tsne2}
\end{figure}

As shown in Tab.~\ref{tab:ana}, MCNL achieves huge improvements compared to Triplet and its two variations.
For example, MCNL outperforms Triplet with $45.2\%$ performance gains in Rank-1 accuracy on Duke and boosts the ReID performance with $11.8\%$ compared to Triplet-same.
It is worth noticing that, compared with Triplet, Triplet-same also improves the ReID performance under the SCT setting.
That is because Triplet-same aims to maximize the distance of the negative pair, of which the two images come from the same camera.
To achieve the above goal, Triplet-same forces the feature extraction model to focus on foreground area and extract more camera-unrelated features because camera-related clues are very similar.
It is obvious that Triplet-other aims to push cross-camera negative pairs away.
Therefore, the model will focus more on background and get worse performance.
Similar to Triplet-same, MCNL also aims to push within-camera negative pairs as far as possible.
Moreover, by restricting the distance of the hardest cross-camera negative pair smaller than the distance of the hardest within-camera negative pair, the model further solves the camera isolation problem and ignores camera-related features.

To better evaluate the above discussions, we utilize t-SNE~\cite{TSNE} to visualize the feature distributions extracted by different methods.
To achieve this, we randomly sample 500 images from the testing set of Duke, and then extract the features on these images through four models trained with Triplet, Triplet-same, Triplet-other, and MCNL, respectively.
Moreover, we use the pseudo \textit{F} statistics~\cite{pf} to evaluate the relations of feature distributions of different cameras quantitatively.
A larger value of pseudo \textit{F} indicates more distinct clusters, which means the extracted features are more related to cameras.
In other words, a smaller value of pseudo \textit{F} implies that features are better learned.

As shown in Fig.~\ref{fig:tsne2}, features extracted by Triplet and Triplet-other are separable according to cameras, which is bad for ReID systems.
Differently, Triplet-same and MCNL both map images to a camera-unrelated feature space.

\begin{figure}[t]
    \begin{center}
        \includegraphics[width=1\linewidth]{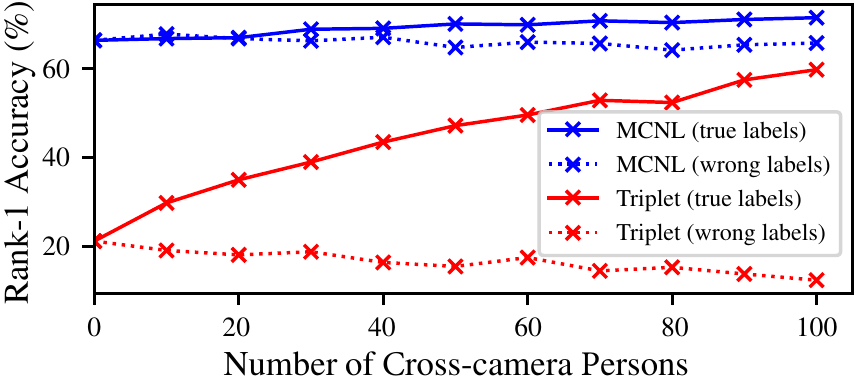}
    \end{center}
    \caption{Rank-1 accuracy (\%) on Duke-SCT with randomly selected cross-camera persons. MCNL shows great robustness against outliers. Solid or dashed line: whether the model receives accurate annotations.
    }
    \label{fig:outlier}
\end{figure}

\textbf{Stability analysis.}
Although the data collection process is restricted, there are inevitably some persons appearing in not only one camera.
To evaluate the robustness of our MCNL under this setting, we conduct experiments on Duke to show how the accuracy changes with respect to the percentage of people showing up in multiple cameras.
As shown in Fig.~\ref{fig:outlier}, when these people are annotated truly according to their identities, Triplet loss benefits largely from ground-truth cross-camera annotations.
Nevertheless, with a considerable portion ($14\%$, $100$ out of $702$) of outliers, MCNL still holds an advantage.
On the other hand, when they are annotated under the SCT setting,~\emph{i.e.}, the images of the same person but different cameras are assigned with different labels, MCNL is quite robust with a small accuracy drop.

This result further demonstrates that our proposed MCNL improves the ReID accuracy under the SCT setting with great robustness against outliers.
In real-world applications, it is easy to control the portion of outliers under a low ratio.

\begin{table*}[htb]
    \centering
    \caption{Comparisons of ReID accuracy (\%) when training with SCT datasets. MCNL reports the best performance on SCT datasets while other methods undergo dramatic accuracy drop. }

    \begin{tabular}{l|c|cc|cc}
        \hline
        \multicolumn{1}{c|}{\multirow{2}[0]{*}{Methods}} & \multicolumn{1}{c|}{\multirow{2}[0]{*}{Ref.}} &  \multicolumn{2}{c|}{Duke-SCT} & \multicolumn{2}{c}{Market-SCT} \\
        \cline{3-6}
        &        & \multicolumn{1}{l}{Rank-1} & \multicolumn{1}{l|}{mAP} & \multicolumn{1}{l}{Rank-1}  & \multicolumn{1}{l}{mAP} \\
        \hline \hline
        Center Loss~\cite{centerloss} & ECCV'16  & 38.7  & 23.2    & 40.3  & 18.5  \\
        A-Softmax~\cite{sphereface} & CVPR'17   & 34.8  & 22.9   & 41.9  & 23.2  \\
        ArcFace~\cite{arcface} & CVPR'19  & 35.8  & 22.8    & 39.4  & 19.8 \\
        \hline
        PCB~\cite{Sun2018_beyond}   & ECCV'18   & 32.7  & 22.2   & 43.5  & 23.5 \\
        Suh's method~\cite{suh_eccv18} & ECCV'18  & 38.5  & 25.4   & 48.0  & 27.3  \\
        MGN~\cite{MGN}   & ACMMM'18 & 27.1  & 18.7  & 38.1  & 24.7  \\
        \hline
        MCNL  & This paper   & \textbf{66.4}  & \textbf{45.3}   & \textbf{66.2}  & \textbf{40.6}  \\
        \hline
    \end{tabular}
    \label{tab:sota}
\end{table*}

\begin{table*}[htb]
    \centering
    \caption{ReID accuracy (\%) comparisons to UT methods. \textit{None} denotes purely unsupervised training without any labels. \textit{Transfer} denotes utilizing other labeled source datasets and unlabeled target datasets. \textit{Tracklet} denotes using tracklet labels. }
    \begin{tabular}{l|c|c|cc|cc}
        \hline
        \multicolumn{1}{c|}{\multirow{2}[0]{*}{Methods}} & \multicolumn{1}{c|}{\multirow{2}[0]{*}{Ref.}}  &
        \multicolumn{1}{c|}{\multirow{2}[0]{*}{Labels}} &   \multicolumn{2}{c|}{Duke} & \multicolumn{2}{c}{Market} \\
        \cline{4-7}
        &       &       & \multicolumn{1}{c}{Rank-1} & \multicolumn{1}{c|}{mAP} & \multicolumn{1}{c}{Rank-1} & \multicolumn{1}{c}{mAP} \\
        \hline \hline
        BOW~\cite{MARKET}   & ICCV'15 & \textit{None}  & 17.1  & 8.3   & 35.8  & 14.8  \\
        DECAMEL~\cite{DECAMEL} & TPAMI'18 & \textit{None}  & - & - & 60.2  & 32.4  \\
        BUC~\cite{Yutian2019}  & AAAI'19 & \textit{None}  & 47.4  & 27.5  & 66.2  & 38.3  \\
        \hline
        TAUDL~\cite{TAUDL} & ECCV'18 & \textit{Tracklet} & 61.7  & 43.5  & 63.7  & 41.2  \\
        \hline
        TJ-AIDL~\cite{TJAIDL} & CVPR'18 & \textit{Transfer} & 44.3  & 23.0  & 58.2  & 26.5  \\
        SPGAN~\cite{SPGAN} & CVPR'18 & \textit{Transfer} & 46.9  & 26.4  & 58.1  & 26.9  \\
        HHL~\cite{HHL}   & ECCV'18 & \textit{Transfer} & 46.9  & 27.2  & 62.2  & 31.4  \\
        MAR~\cite{MAR}   & CVPR'19 & \textit{Transfer} & {67.1 } & {48.0 } & 67.7  & 40.0  \\
        ECN~\cite{ECN}   & CVPR'19 & \textit{Transfer} & 63.3  & 40.4  & {75.1 } & {43.0 } \\
        \hline
        MCNL  & This paper & SCT   & 66.4  & 45.3  & 66.2  & 40.6  \\
        MCNL+MAR~\cite{MAR} & This paper & \textit{Transfer}+SCT & \textbf{71.4} &	\textbf{53.3} &	72.3 &	48.0 \\
        MCNL+ECN~\cite{ECN} & This paper & \textit{Transfer}+SCT & 67.3 &	45.5 &	\textbf{76.3} &	\textbf{51.2}  \\
        \hline
    \end{tabular}
    \label{tab:state}
\end{table*}

\subsection{Comparison to Previous Work} \label{sec:5.4}
\textbf{Comparisons to FST methods.}
We evaluate a few popular FST methods under the SCT setting and compare our method with other advanced metric learning algorithms.
As shown in Tab.~\ref{tab:sota}, previous state-of-the-art methods for the FST setting fail dramatically under the SCT setting while MCNL shows great advantages.
This is because, without cross-camera annotations, these methods are unable to extract camera-unrelated features.

\textbf{Comparisons to UT methods.}
Our motivation of the SCT setting is for fast deployment of ReID systems on new target scenes, which is the same as the motivation of the UT setting.
Thus, as shown in Tab.~\ref{tab:state}, we also compare MCNL with previous unsupervised training methods, including purely unsupervised methods, tracklet association learning method, and domain adaptation methods.

Compared to the state-of-the-art purely unsupervised methods (labels denoted as \textit{None}), our proposed MCNL significantly outperforms BUC~\cite{Yutian2019} with $19.0\%$ performance gains in Rank-1 accuracy on Duke.
As for TAUDL~\cite{TAUDL} that uses \textit{Tracklet} labels, the entire training sets are used to train the models.
Our method constructs SCT datasets for training, and thus only a small portion of training data are used, but still surpasses TAUDL on Duke and Market in Rank-1 accuracy.

Recently, many domain adaptation methods that use other labeled datasets for extra supervision obtain good ReID accuracy.
Our MCNL alone achieves competitive results compared to them.
Moreover, our method is also complementary to current domain adaptation methods and can be easily combined by replacing the target datasets with SCT datasets.
Such combination instantly brings significant improvement.
We take MAR~\cite{MAR} and ECN~\cite{ECN}, for example.
After using SCT data and MCNL in MAR, we boost $4.3\%$ in Rank-1 accuracy and $5.3\%$ in mAP on Duke dataset; the combination of MCNL and ECN improves mAP on Market by $8.2\%$ compared to ECN only.
Taking the advantages of reliable target domain annotations and extra transferred information, we achieve the best ReID performance on Duke and Market, respectively.
Implementation details about the combination of MCNL and MAR/ECN will be included in the supplemental materials.

Note that, our method achieves good performance with much fewer training data.
Because of giving up collecting cross-camera pedestrian images under the SCT setting, the above training data can be easily collected and annotated.
Therefore, compared with prior work, our method and proposed SCT setting are more suitable for fast deployment of ReID systems with good performance on new target scenes.

\section{Conclusions}
In this paper, we explore a new setting named single-camera-training (SCT) for person ReID. With the advantage of low costs in data collection and annotation, SCT lays the foundation of fast deployment of ReID systems in new environments. To work under SCT, we propose a novel loss term named multi-camera negative loss (MCNL). Experiments demonstrate that under SCT, the proposed approach boosts ReID performance of existing approaches by a large margin.

Our approach reveals the possibility of learning cross-camera correspondence without cross-camera annotations.
In the future, we will explore more cues to leverage under the SCT setting and consider the mixture of single-camera and cross-camera annotations to further improve ReID accuracy.

\newpage
{\small
\bibliographystyle{aaai}
\bibliography{egbib}

\begin{thebibliography}{}

\bibitem[\protect\citeauthoryear{Cali{\'n}ski and Harabasz}{1974}]{pf}
Cali{\'n}ski, T., and Harabasz, J.
\newblock 1974.
\newblock A dendrite method for cluster analysis.
\newblock {\em Communications in Statistics-theory and Methods} 3(1):1--27.

\bibitem[\protect\citeauthoryear{Chen \bgroup et al\mbox.\egroup
  }{2017}]{Quadruplet}
Chen, W.; Chen, X.; Zhang, J.; and Huang, K.
\newblock 2017.
\newblock Beyond triplet loss: a deep quadruplet network for person
  re-identification.
\newblock In {\em CVPR}.

\bibitem[\protect\citeauthoryear{Deng \bgroup et al\mbox.\egroup
  }{2009}]{ImageNet}
Deng, J.; Dong, W.; Socher, R.; Li, L.-J.; Li, K.; and Fei-Fei, L.
\newblock 2009.
\newblock Imagenet: A large-scale hierarchical image database.
\newblock In {\em CVPR}.

\bibitem[\protect\citeauthoryear{Deng \bgroup et al\mbox.\egroup
  }{2018}]{SPGAN}
Deng, W.; Zheng, L.; Kang, G.; Yang, Y.; Ye, Q.; and Jiao, J.
\newblock 2018.
\newblock Image-image domain adaptation with preserved self-similarity and
  domain-dissimilarity for person reidentification.
\newblock In {\em CVPR}.

\bibitem[\protect\citeauthoryear{Deng \bgroup et al\mbox.\egroup
  }{2019}]{arcface}
Deng, J.; Guo, J.; Niannan, X.; and Zafeiriou, S.
\newblock 2019.
\newblock Arcface: Additive angular margin loss for deep face recognition.
\newblock In {\em CVPR}.

\bibitem[\protect\citeauthoryear{Gray and Tao}{2008}]{ELF}
Gray, D., and Tao, H.
\newblock 2008.
\newblock Viewpoint invariant pedestrian recognition with an ensemble of
  localized features.
\newblock In {\em ECCV}.

\bibitem[\protect\citeauthoryear{He \bgroup et al\mbox.\egroup }{2016}]{ResNet}
He, K.; Zhang, X.; Ren, S.; and Sun, J.
\newblock 2016.
\newblock Deep residual learning for image recognition.
\newblock In {\em CVPR}.

\bibitem[\protect\citeauthoryear{Hermans, Beyer, and Leibe}{2017}]{InDefense}
Hermans, A.; Beyer, L.; and Leibe, B.
\newblock 2017.
\newblock In defense of the triplet loss for person re-identification.
\newblock {\em arXiv preprint arXiv:1703.07737}.

\bibitem[\protect\citeauthoryear{Keuper \bgroup et al\mbox.\egroup
  }{2018}]{tracking}
Keuper, M.; Tang, S.; Andres, B.; Brox, T.; and Schiele, B.
\newblock 2018.
\newblock Motion segmentation \& multiple object tracking by correlation
  co-clustering.
\newblock {\em IEEE transactions on pattern analysis and machine intelligence}.

\bibitem[\protect\citeauthoryear{Kingma and Ba}{2014}]{ADAM}
Kingma, D.~P., and Ba, J.
\newblock 2014.
\newblock Adam: A method for stochastic optimization.
\newblock {\em arXiv preprint arXiv:1412.6980}.

\bibitem[\protect\citeauthoryear{Li, Zhu, and Gong}{2018}]{TAUDL}
Li, M.; Zhu, X.; and Gong, S.
\newblock 2018.
\newblock Unsupervised person re-identification by deep learning tracklet
  association.
\newblock In {\em ECCV}.

\bibitem[\protect\citeauthoryear{Liang \bgroup et al\mbox.\egroup
  }{2015}]{Liang2015}
Liang, C.; Huang, B.; Hu, R.; Zhang, C.; Jing, X.; and Xiao, J.
\newblock 2015.
\newblock A unsupervised person re-identification method using model based
  representation and ranking.
\newblock In {\em ACM MM}.

\bibitem[\protect\citeauthoryear{Liao \bgroup et al\mbox.\egroup }{2015}]{LOMO}
Liao, S.; Hu, Y.; Zhu, X.; and Li, S.~Z.
\newblock 2015.
\newblock Person re-identification by local maximal occurrence representation
  and metric learning.
\newblock In {\em CVPR}.

\bibitem[\protect\citeauthoryear{Lin \bgroup et al\mbox.\egroup
  }{2019}]{Yutian2019}
Lin, Y.; Dong, X.; Zheng, L.; Yan, Y.; and Yang, Y.
\newblock 2019.
\newblock A bottom-up clustering approach to unsupervised person
  re-identification.
\newblock In {\em AAAI}.

\bibitem[\protect\citeauthoryear{Liu \bgroup et al\mbox.\egroup
  }{2017}]{sphereface}
Liu, W.; Wen, Y.; Yu, Z.; Li, M.; Raj, B.; and Song, L.
\newblock 2017.
\newblock Sphereface: Deep hypersphere embedding for face recognition.
\newblock In {\em CVPR}.

\bibitem[\protect\citeauthoryear{Liu \bgroup et al\mbox.\egroup
  }{2018}]{POSETRANS}
Liu, J.; Ni, B.; Yan, Y.; Zhou, P.; Cheng, S.; and Hu, J.
\newblock 2018.
\newblock Pose transferrable person re-identification.
\newblock In {\em CVPR}.

\bibitem[\protect\citeauthoryear{Luo \bgroup et al\mbox.\egroup
  }{2019}]{tracking2}
Luo, W.; Stenger, B.; Zhao, X.; and Kim, T.-K.
\newblock 2019.
\newblock Trajectories as topics: Multi-object tracking by topic discovery.
\newblock {\em IEEE Transactions on Image Processing} 28(1):240--252.

\bibitem[\protect\citeauthoryear{Peng \bgroup et al\mbox.\egroup
  }{2016}]{peng2016}
Peng, P.; Xiang, T.; Wang, Y.; Pontil, M.; Gong, S.; Huang, T.; and Tian, Y.
\newblock 2016.
\newblock Unsupervised cross-dataset transfer learning for person
  re-identification.
\newblock In {\em CVPR}.

\bibitem[\protect\citeauthoryear{Schroff, Kalenichenko, and
  Philbin}{2015}]{FACENET}
Schroff, F.; Kalenichenko, D.; and Philbin, J.
\newblock 2015.
\newblock Facenet: A unified embedding for face recognition and clustering.
\newblock In {\em CVPR}.

\bibitem[\protect\citeauthoryear{Shi \bgroup et al\mbox.\egroup
  }{2016}]{Moderate}
Shi, H.; Yang, Y.; Zhu, X.; Liao, S.; Lei, Z.; Zheng, W.; and Li, S.~Z.
\newblock 2016.
\newblock Embedding deep metric for person re-identification: A study against
  large variations.
\newblock In {\em ECCV}.

\bibitem[\protect\citeauthoryear{Suh \bgroup et al\mbox.\egroup
  }{2018}]{suh_eccv18}
Suh, Y.; Wang, J.; Tang, S.; Mei, T.; and Lee, K.~M.
\newblock 2018.
\newblock Part-aligned bilinear representations for person re-identification.
\newblock In {\em ECCV}.

\bibitem[\protect\citeauthoryear{Sun \bgroup et al\mbox.\egroup
  }{2017}]{SVDNET}
Sun, Y.; Zheng, L.; Deng, W.; and Wang, S.
\newblock 2017.
\newblock Svdnet for pedestrian retrieval.
\newblock In {\em ICCV}.

\bibitem[\protect\citeauthoryear{Sun \bgroup et al\mbox.\egroup
  }{2018}]{Sun2018_beyond}
Sun, Y.; Zheng, L.; Yang, Y.; Tian, Q.; and Wang, S.
\newblock 2018.
\newblock Beyond part models: Person retrieval with refined part pooling (and a
  strong convolutional baseline).
\newblock In {\em ECCV}.

\bibitem[\protect\citeauthoryear{Van Der~Maaten}{2014}]{TSNE}
Van Der~Maaten, L.
\newblock 2014.
\newblock Accelerating t-sne using tree-based algorithms.
\newblock {\em The Journal of Machine Learning Research} 15(1):3221--3245.

\bibitem[\protect\citeauthoryear{Wang \bgroup et al\mbox.\egroup }{2018a}]{MGN}
Wang, G.; Yuan, Y.; Chen, X.; Li, J.; and Zhou, X.
\newblock 2018a.
\newblock Learning discriminative features with multiple granularities for
  person re-identification.
\newblock In {\em ACM MM}.

\bibitem[\protect\citeauthoryear{Wang \bgroup et al\mbox.\egroup
  }{2018b}]{TJAIDL}
Wang, J.; Zhu, X.; Gong, S.; and Li, W.
\newblock 2018b.
\newblock Transferable joint attribute-identity deep learning for unsupervised
  person re-identification.
\newblock In {\em CVPR}.

\bibitem[\protect\citeauthoryear{Wang \bgroup et al\mbox.\egroup
  }{2019}]{streid}
Wang, G.; Lai, J.; Huang, P.; and Xie, X.
\newblock 2019.
\newblock Spatial-temporal person re-identification.
\newblock In {\em AAAI}.

\bibitem[\protect\citeauthoryear{Wei \bgroup et al\mbox.\egroup }{2017}]{GLAD}
Wei, L.; Zhang, S.; Yao, H.; Gao, W.; and Tian, Q.
\newblock 2017.
\newblock Glad: global-local-alignment descriptor for pedestrian retrieval.
\newblock In {\em ACM MM}.

\bibitem[\protect\citeauthoryear{Wei \bgroup et al\mbox.\egroup }{2018}]{PTGAN}
Wei, L.; Zhang, S.; Gao, W.; and Tian, Q.
\newblock 2018.
\newblock Person transfer gan to bridge domain gap for person
  re-identification.
\newblock In {\em CVPR}.

\bibitem[\protect\citeauthoryear{Wen \bgroup et al\mbox.\egroup
  }{2016}]{centerloss}
Wen, Y.; Zhang, K.; Li, Z.; and Qiao, Y.
\newblock 2016.
\newblock A discriminative feature learning approach for deep face recognition.
\newblock In {\em ECCV}.
\newblock Springer.

\bibitem[\protect\citeauthoryear{Yu \bgroup et al\mbox.\egroup }{2019}]{MAR}
Yu, H.-X.; Zheng, W.-S.; Wu, A.; Guo, X.; Gong, S.; and Lai, J.-H.
\newblock 2019.
\newblock Unsupervised person re-identification by soft multilabel learning.
\newblock In {\em CVPR}.

\bibitem[\protect\citeauthoryear{{Yu}, {Wu}, and {Zheng}}{2018}]{DECAMEL}
{Yu}, H.; {Wu}, A.; and {Zheng}, W.
\newblock 2018.
\newblock Unsupervised person re-identification by deep asymmetric metric
  embedding.
\newblock {\em IEEE Transactions on Pattern Analysis and Machine Intelligence}.

\bibitem[\protect\citeauthoryear{Zhang \bgroup et al\mbox.\egroup
  }{2017}]{alignedreid}
Zhang, X.; Luo, H.; Fan, X.; Xiang, W.; Sun, Y.; Xiao, Q.; Jiang, W.; Zhang,
  C.; and Sun, J.
\newblock 2017.
\newblock Alignedreid: Surpassing human-level performance in person
  re-identification.
\newblock {\em arXiv preprint arXiv:1711.08184}.

\bibitem[\protect\citeauthoryear{Zheng \bgroup et al\mbox.\egroup
  }{2015}]{MARKET}
Zheng, L.; Shen, L.; Tian, L.; Wang, S.; Wang, J.; and Tian, Q.
\newblock 2015.
\newblock Scalable person re-identification: A benchmark.
\newblock In {\em ICCV}.

\bibitem[\protect\citeauthoryear{Zheng, Karanam, and Radke}{2018}]{RPIfield}
Zheng, M.; Karanam, S.; and Radke, R.~J.
\newblock 2018.
\newblock Rpifield: A new dataset for temporally evaluating person
  re-identification.
\newblock In {\em CVPR Workshops}.

\bibitem[\protect\citeauthoryear{Zheng, Yang, and
  Hauptmann}{2016}]{zheng2016person}
Zheng, L.; Yang, Y.; and Hauptmann, A.~G.
\newblock 2016.
\newblock Person re-identification: Past, present and future.
\newblock {\em arXiv preprint arXiv:1610.02984}.

\bibitem[\protect\citeauthoryear{Zheng, Zheng, and Yang}{2017a}]{zheng2016a}
Zheng, Z.; Zheng, L.; and Yang, Y.
\newblock 2017a.
\newblock A discriminatively learned cnn embedding for person reidentification.
\newblock {\em ACM Transactions on Multimedia Computing, Communications, and
  Applications} 14(1):13.

\bibitem[\protect\citeauthoryear{Zheng, Zheng, and Yang}{2017b}]{DUKEMTMC}
Zheng, Z.; Zheng, L.; and Yang, Y.
\newblock 2017b.
\newblock Unlabeled samples generated by gan improve the person
  re-identification baseline in vitro.
\newblock In {\em ICCV}.

\bibitem[\protect\citeauthoryear{Zheng, Zheng, and Yang}{2018}]{PAN}
Zheng, Z.; Zheng, L.; and Yang, Y.
\newblock 2018.
\newblock Pedestrian alignment network for large-scale person
  re-identification.
\newblock {\em IEEE Transactions on Circuits and Systems for Video Technology}.

\bibitem[\protect\citeauthoryear{Zhong \bgroup et al\mbox.\egroup }{2018}]{HHL}
Zhong, Z.; Zheng, L.; Li, S.; and Yang, Y.
\newblock 2018.
\newblock Generalizing a person retrieval model hetero-and homogeneously.
\newblock In {\em ECCV}.

\bibitem[\protect\citeauthoryear{Zhong \bgroup et al\mbox.\egroup }{2019}]{ECN}
Zhong, Z.; Zheng, L.; Luo, Z.; Li, S.; and Yang, Y.
\newblock 2019.
\newblock Invariance matters: Exemplar memory for domain adaptive person
  re-identification.
\newblock In {\em CVPR}.

\end{thebibliography}
}
\newpage
\begin{figure*}[!b]
    \centering
    \includegraphics[width=1.0\textwidth]{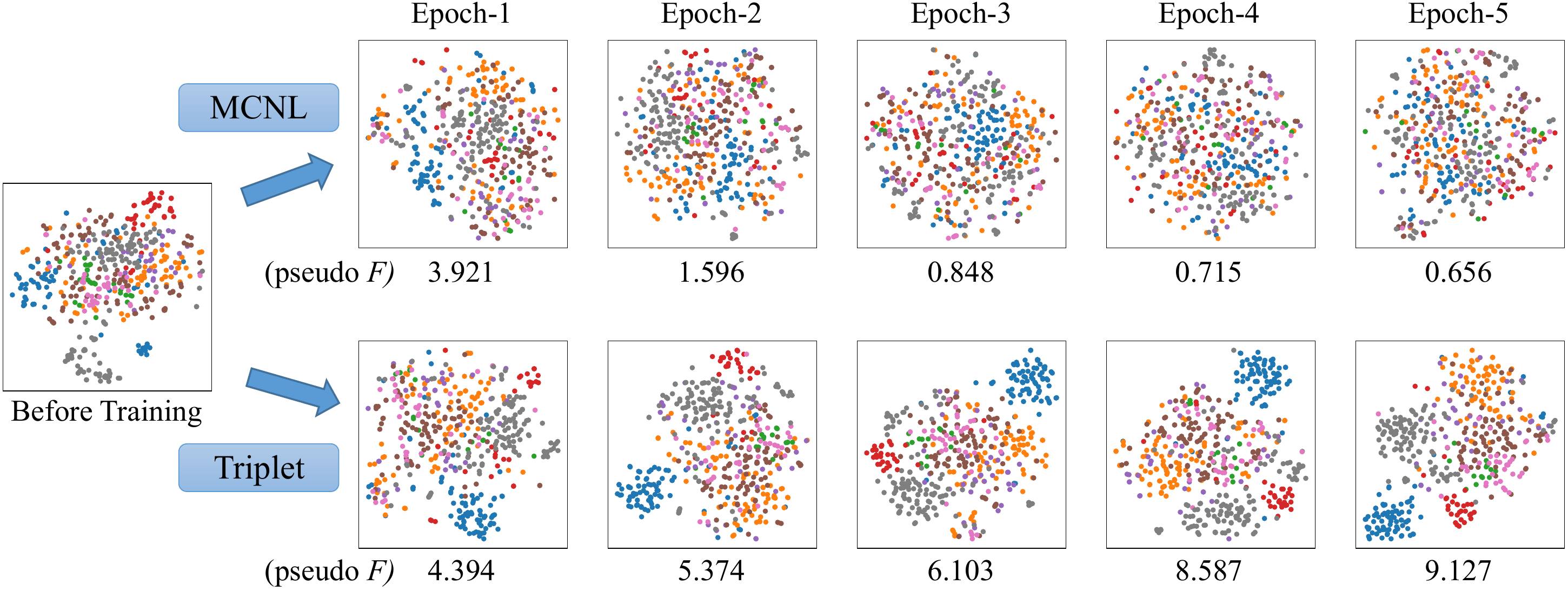}
    \caption{Visualization of learned feature distributions at the early training stage. Pseudo \textit{F} statistics are shown below each sub-figure. The sub-figures in the top row are produced by models trained with MCNL, and those in the bottom row are produced by models trained with Triplet. The leftmost sub-figure is produced by the pre-trained model on ImageNet. Each color indicates features from a camera. This figure is best viewed in color.}
    \label{fig:verybig}
\end{figure*}
\section*{Supplemental Material}
\subsection*{A. Implementation Details of MCNL+MAR/ECN}
\textbf{MCNL+MAR.}
MAR~\cite{MAR} utilizes a large-scale reference dataset (MSMT17~\cite{PTGAN}) to learn soft multilabels for unlabeled target datasets.
We first replace the target datasets with their SCT versions,~\emph{i.e.}, Duke-SCT and Market-SCT.
Then, we add MCNL to the loss function with a hyperparameter $\lambda_3$ to balance the values of losses.
The original loss function of MAR is:
\begin{equation}
  \mathcal{L}_\mathrm{MAR}=\mathcal{L}_\mathrm{MDL}+\lambda_1\mathcal{L}_\mathrm{CML}+\lambda_2\mathcal{L}_\mathrm{RAL},
\end{equation}
and the loss function of MCNL+MAR is:
\begin{equation}
  \mathcal{L}_\mathrm{MAR}=\mathcal{L}_\mathrm{MDL}+\lambda_1\mathcal{L}_\mathrm{CML}+\lambda_2\mathcal{L}_\mathrm{RAL}+\lambda_3\mathcal{L}_\mathrm{MCNL}.
\end{equation}
We keep all hyperparameters as same as MAR, and set $\lambda_3=10$ for Market-SCT and  $\lambda_3=20$ for Duke-SCT.
Apart from the target training data and loss function, nothing has been changed.
\textbf{MCNL+ECN.}
ECN~\cite{ECN} discovers three invariance factors for unsupervised person ReID.
We first replace the target training data with SCT versions.
During training, ECN uses a GAN model to generate fake images with different camera styles.
We only use those fake images whose real images are in our sampled SCT datasets.
Then, we add MCNL to the loss function.
The original loss function of ECN is:
\begin{equation}
  \mathcal{L}_\mathrm{ECN}=(1-\lambda)\mathcal{L}_\mathrm{src}+\lambda\mathcal{L}_\mathrm{tgt},
\end{equation}
and after we add MCNL, the loss function of MCNL+ECN is:
\begin{equation}
  \mathcal{L}_\mathrm{ECN}=(1-\lambda)\mathcal{L}_\mathrm{src}+\lambda(\mathcal{L}_\mathrm{tgt}+\lambda_2\mathcal{L}_\mathrm{MCNL}).
\end{equation}
We set $\lambda_2=10$ and keep other hyperparameters the same as ECN.
Moreover, after introducing the SCT data with reliable labels, the exemplar memory of ECN now has key-value memory according to the identity labels.
The original purpose of exemplar-invariance is to classify each image into its own class.
After the combination, exemplar-invariance loss classifies each identity.
Since exemplar-invariance loss now pulls positive pairs close, neighborhood-invariance loss is no longer needed.
We change exemplar-invariance loss accordingly and remove neighborhood-invariance loss.
\subsection*{B. Visualization of Feature Distributions}
We also make a comparison between MCNL and Triplet to see the difference during the training process.
From Fig.~\ref{fig:verybig}, we can conclude that at the very beginning of training, Triplet learns more camera-related features while MCNL quickly discards camera information and learns more person-related features.
The observation further demonstrates MCNL can alleviate the camera isolation problem, and make a uniform feature space for all cameras.

\end{document}